# Fully Automated Photogrammetric Data Segmentation and Object Information Extraction Approach for Creating Simulation Terrain

**Meida Chen, Andrew Feng, Kyle McCullough, Pratusha Bhuvana Prasad, Ryan McAlinden**
*USC Institute for Creative Technologies*
**Los Angeles, California**
{mechen, feng, McCullough, bprasad, mcalinden}@ict.usc.edu

**Lucio Soibelman**
*USC Department of Civil and Environmental Engineering*
**Los Angeles, California**
soibelman@usc.edu

**Mike Enloe**
*Synthetic Training Environment Cross Functional Team*
**Orlando, Florida**
Michael.r.enloe.civ@mail.mil

## ABSTRACT

Our previous works have demonstrated that visually realistic 3D meshes can be automatically reconstructed with low-cost, off-the-shelf unmanned aerial systems (UAS) equipped with capable cameras, and efficient photogrammetric software techniques (McAlinden, Suma, Grechkin, & Enloe, 2015; Spicer, McAlinden, Conover, & Adelphi, 2016). However, such generated data do not contain semantic information/features of objects (i.e., man-made objects, vegetation, ground, object materials, etc.) and cannot allow the sophisticated user-level and system-level interaction. Considering the use case of the data in creating realistic virtual environments for training and simulations (i.e., mission planning, rehearsal, threat detection, etc.), segmenting the data and extracting object information are essential tasks. Previous studies have focused on and made valuable contributions to segment Light Detection and Ranging (LIDAR) generated 3D point clouds and classifying ground materials from real-world images. However, only a few studies have focused on the data created using the photogrammetric technique.

Thus, the objective of this research is to design and develop a fully automated photogrammetric data segmentation and object information extraction framework. This designed framework utilizes concepts from the areas of computer vision and deep learning. To validate the proposed framework, the segmented data and extracted features were used to create virtual environments in the authors previously designed simulation tool—i.e., Aerial Terrain Line of Sight Analysis System (ATLAS). The results showed that 3D mesh trees could be replaced with geo-typical 3D tree models using the extracted individual tree locations. The extracted tree features (i.e., color, width, height) are valuable for selecting the appropriate tree species and enhance visual quality. Furthermore, the identified ground material information can be taken into consideration for pathfinding. The shortest path can be computed not only considering the physical distance, but also considering the off-road vehicle performance capabilities on different ground surface materials.

## ABOUT THE AUTHORS

**Meida Chen** is currently a research assistant at the University of Southern California's Institute for Creative Technologies (USC-ICT) working on One World Terrain project. He is pursuing his Ph.D. degree at USC Sonny Astani Department of Civil and Environmental Engineering. His research focuses on the semantic modeling of outdoor scenes for the creation of virtual environments and simulations. Email: mechen@ict.usc.edu

**Andrew Feng** is currently a research scientist at USC-ICT working on One World Terrain project. Previously, he was a research associate focusing on character animation and automatic 3D avatar generation. His research work involves applying machine learning techniques to solve computer graphics problems such as animation synthesis, mesh skinning, and mesh deformation. He received the Ph.D. and MS degree in computer science from University of Illinois at Urbana-Champaign. Email: feng@ict.usc.edu

**Kyle McCullough** is currently the lead Programmer at USC-ICT working on the One World Terrain project. Previously, he was a creative director and writer for the Video Games Industry, most recently for Ubisoft's 'Transference', winning Best Interactive Narrative VR Experience at Raindance 2018. His research work involves advanced prototype systems development, utilizing AI and 3D visualization to increase fidelity and realism in large-scale dynamic simulation environments. He has a B.F.A. from New York University. Email: McCullough@ict.usc.edu

**Pratusha Bhuvana Prasad** is currently a researcher at USC-ICT working on One World Terrain project. Her research focuses on computer vision for geometry and using machine learning methods to solve the same. She has a master's degree from Ming Hsieh Department of Electrical and Computer Engineering, USC.
Email: bprasad@ict.usc.edu

**Ryan McAlinden** is the Associate Director for Digital Training and Instruction at USC-ICT. He rejoined ICT in 2013 after a three-year post as a senior scientist at the NATO Communications & Information Agency (NCIA) in The Hague, Netherlands. There he led the provision of operational analysis support to the International Security Assistance Force (ISAF) Headquarters in Kabul, Afghanistan. Prior to joining NCIA, Ryan worked as a computer scientist at USC-ICT from 2002 through 2009. He has a B.S. from Rutgers University and M.S. in computer science from USC. Email: mcalinden@ict.usc.edu

**Lucio Soibelman** is a Professor and Chair of the Sonny Astani Department of Civil and Environmental Engineering at USC. Dr. Soibelman's research focuses on use of information technology for economic development, information technology support for construction management, process integration during the development of large-scale engineering systems, information logistics, artificial intelligence, data mining, knowledge discovery, image reasoning, text mining, machine learning, multi-reasoning mechanisms, sensors, sensor networks, and advanced infrastructure systems. Email: soibelman@usc.edu

**Mike Enloe** is the Chief Engineer and an originator of the Synthetic Training Environment (STE) Cross Functional Team (CFT) for the US Army located in Orlando, Fl. Mike has been a Government Civilian in his current position for 8 years, and has over 15 years of experience in modeling and simulation. Mike's duty is to serve as the lead technical consultant to the 2-Star General Office CFT Director and to steer & oversee science and technology research that will help mitigate gaps in the design of the Army's Synthetic Training Environment. Mike has supported the engineering of many Army training simulation systems, to include the linkage of Mission Command Systems to computer simulations, mission planning/rehearsal tools, low overhead battle staff trainers and Army Games for Training. Mike has a bachelor's degree in Computer Science from Phoenix University. Email: michael.r.enloe.civ@mail.mil

# Fully Automated Photogrammetric Data Segmentation and Object Information Extraction Approach for Creating Simulation Terrain

| | | |
|---|---|---|
| **Meida Chen, Andrew Feng, Kyle McCullough, Pratusha Bhuvana Prasad, Ryan McAlinden**<br>*USC Institute for Creative Technologies*<br>Los Angeles, California<br>{mechen, feng, McCullough, bprasad, mcalinden}@ict.usc.edu | **Lucio Soibelman**<br>*USC Department of Civil and Environmental Engineering*<br>Los Angeles, California<br>soibelman@usc.edu | **Mike Enloe**<br>*Synthetic Training Environment Cross Functional Team*<br>Orlando, Florida<br>Michael.r.enloe.civ@mail.mil |

## INTRODUCTION

Photogrammetric techniques have dramatically improved over the past few years, enabling the creation of visually compelling 3D meshes using unmanned aerial vehicle imagery. These high-quality 3D meshes have attracted notice from both academicians and industry practitioners in developing virtual environments and simulations. However, the photogrammetric-generated point clouds/meshes do not allow both user- and system-level interaction, as they do not contain the semantic information to distinguish between objects.

Consider the case of training soldiers in a virtual environment with 3D meshes representing the scene. The task is to recognize the shortest path from location A to location B in which the individual is visible from a given vantage point. With artificial intelligence (AI) search algorithms, the shortest path could be computed, and penalties could be assigned to a route based on the number of obstructions blocking the enemy's line-of-sight. However, in reality, line-of-sight that is blocked by concrete walls, glass windows, and trees should be assigned different penalties when considering a route, since some materials cannot protect soldiers from gunshots (i.e., glass windows).

Being able to segment, classify, and recognize distinct types of objects together, along with identifying and extracting associated features (i.e., individual tree locations and ground materials) in the generated meshes, are essential tasks in creating realistic virtual simulations. Rendering different objects in a virtual environment and assigning actual physical properties to each of them will not only enhance the visual quality but also allow various user interactions with a terrain model. An additional example would be computing the trafficability of an entity in the virtual environment; terrain surfaces must be classified properly based on their material composition since different terrain surface materials such as bare soil, grass, rock, mud, etc. could affect the off-road vehicle performance (i.e., the speed of driving a vehicle on grass is different from driving a vehicle on bare soil). Though these examples are an oversimplification, they emphasize the point that, without semantic segmentation of the mesh data, realistic virtual simulations cannot be achieved.

In this paper, a fully automated point clouds/meshes segmentation and object information extraction framework is designed and developed to support next-generation modeling, simulation & training. The extracted semantic information - i.e., top-level terrain elements (ground, man-made structures, vegetations), ground materials (paved roads, bare earth, grass), and individual tree locations will enable more realistic simulation with additional terrain parameters such as maneuvering capability analysis, maximum standoff distance, etc. While the previous works exist on processing 2D aerial imagery or LiDAR points, in this paper we focus on segmentation for photogrammetric-generated data. The main challenges include training on noisy data from photogrammetric reconstruction, annotating a large amount of 2D/3D data, and generalizing the trained model across different geographic regions. In this work, modern machine learning techniques - i.e., 3D convolutional neural networks (CNN) were utilized to tackle these challenges.

It is also worth pointing out that the presented study in this paper is part of the One World Terrain (OWT) project. One of the objectives of the OWT project is to provide small units with the organic capability to create geo-specific virtual environments for training and rehearsal purposes to support military operations. For more information about the OWT project, readers can refer to http://ict.usc.edu/prototypes/one-world-terrain-owt/.

## SEMANTIC TERRAIN POINT LABELING SYSTEM PLUS - STPLS+

This research aims to investigate and develop a framework for the segmentation/classification of photogrammetric generated point cloud into predefined categories and extract object information for the creation of virtual environments and simulations. The methodology combines concepts from the areas of computer vision and machine learning. Figure 1 presents the designed STPLS+ framework that illustrates the workflow, emphasizing the main elements and steps involved in the process. Ground materials are identified using the orthophotos; the hypothesis behind this research is that materials can be identified based on the 2D textures/patterns instead of 3D geometric shapes. Point clouds are segmented into top-level terrain elements (i.e., ground, man-made objects, and vegetation) using the deep learning algorithm (i.e., U-net). Tree locations are identified from the segmented tree points; the related features for each tree (i.e., width, height, and average color) are also extracted. Finally, since 3D meshes are needed instead of point clouds in modern game engines to enable 3D interaction and collision detection, meshes are segmented based on the point cloud segmentation results.

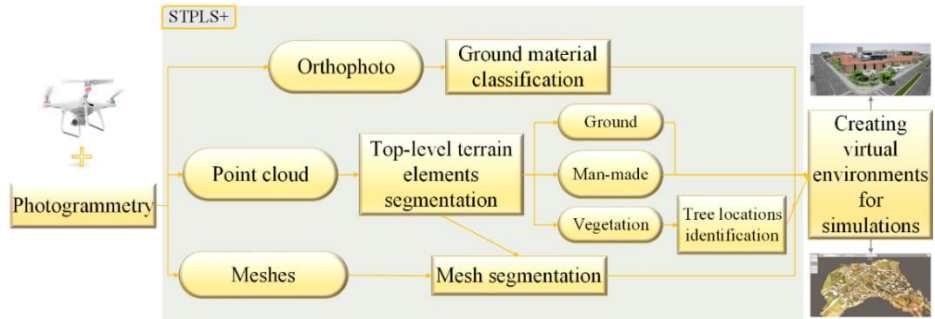

**Figure 1. Workflow of STPLS+**

**Top-Level Terrain Elements Segmentation**

The authors' previous work (Chen, McAlinden, Spicer, & Soibelman, 2019) has demonstrated the capability of using handcrafted point features (i.e., color-based, point density-based, and local surface-based features) with supervised machine learning algorithms (i.e., Support Vector Machine and Random Forest) for photogrammetric-generated point cloud segmentation. One limitation of using handcrafted point features is that a pre-trained segmentation model using existing datasets cannot be applied/reused on a newly collected dataset. This is because point clouds with different quality may yield different values for the handcrafted point features and the quality of a photogrammetric-generated point cloud highly depends on the parameters used for aerial photo collection (i.e., flight altitude and overlap between images). In practice, these parameter settings are defined based on the maximum allowable time for data collection, the workforce talent available, and the equipment available. Consequently, our previous developed point cloud segmentation system can only be used in a semi-automatic fashion where training data needs to be manually created every time new point clouds are generated.

To overcome such a limitation, more robust features need to be developed and a generalizable segmentation model needs to be trained with existing datasets. The state-of-the-art deep learning techniques - i.e., Deep Neural Networks (DNN) architectures provide the suitable initial foundation to build the automatic pipeline. Previous works have applied DNN successfully on segmenting 3D data such as outdoor LiDAR point clouds (Landrieu & Simonovsky, 2018). However, no research works have been done in applying DNN for semantic segmentation of photogrammetric-generated terrain.

Since our datasets contain large scale point clouds across different geographic regions, it poses additional challenges to train a model that generalizes well for different maps. Furthermore, photogrammetric-generated point clouds tend to be noisy—in some case ground cannot be captured due to dense canopy from trees, and vegetation tends to appear as solid blobs instead of individual tree with well-formed branches (Spicer, McAlinden, & Conover, 2016)—which makes our segmentation task even more challenging than working with the LiDAR data. The designed fully automatic pipeline for top-level terrain elements segmentation is shown in Figure 2. Data preprocessing is needed to remove noises and select the area of interest from the raw point clouds. Following that, voxel grids are generated based on the cleaned point cloud and will be used for segmentation. U-Net is selected in this study to segment the point cloud into the ground, man-made objects, and vegetation. Details of the designed point cloud segmentation workflow are discussed in the following sections.

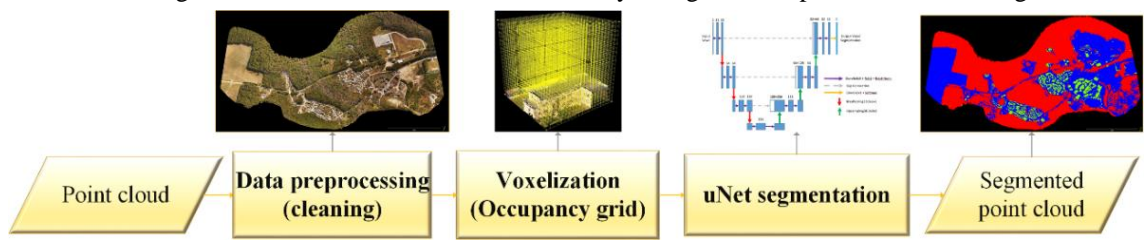

**Figure 2. Top-level Terrain Elements Segmentation**

**Data preprocessing**

Photogrammetric-generated point clouds tend to be noisy. Artifacts appear under the ground due to the limitation of the used photogrammetric software (i.e., ContextCapture). Underground noise and artifacts are shown in Figure 3 (a). These artifacts can dramatically affect the segmentation results since they appear with random shapes, sizes, and at random locations. From an algorithmic point of view, a segmentation model learns to label a point as ground, if its position is relatively low compared to its surrounding points. This will not be the case if there are noisy points (artifacts) under the terrain.

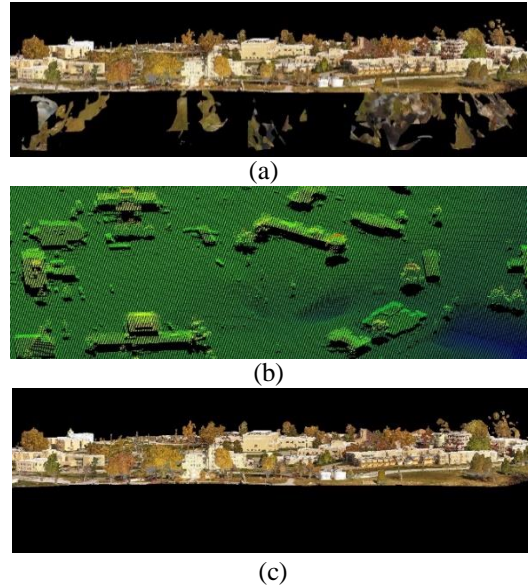

To remove these artifacts, we used the digital surface model (DSM) in a point cloud format which can also be generated from the photogrammetric software. A point *p* in the point cloud is considered as noise, if the z coordinate of *p* is smaller than the lowest point of *p*'s neighbor points from the DSM. Since the DSM does not contain points on the vertical surfaces as shown in Figure 3 (b), simply searching neighbor points in a sphere will also remove points on vertical surfaces (i.e., points on walls). Points in DSM are considered as *p*'s neighbor points if they are within a cylinder of a radius r centered at p, where r is a predefined threshold. If the *r* value is too small, points under a tree or any upside-down cone-shaped object will be removed. If the *r* value is too big, underground noises in a hilly terrain cannot be removed. The r value is set to 5 meters in this study. The cleaned point cloud is shown in Figure 3 (c).

**Figure 3. Data Cleaning. (a) Point Cloud with Artifacts. (b) Digital Surface Model (DSM). (c) Cleaned Point Cloud.**

Setting the camera orientation to 45 degrees forward (oblique) is preferred for collecting images to overcome the challenges for reconstructing vertical surfaces in prior work (McAlinden et al., 2015). However, with 45 degrees forward camera orientation, the collected images cover more areas (unwanted areas) than the predefined Area of Interest (AOI). After the photogrammetric 3D reconstruction process, the unwanted areas are also reconstructed with images that do not have enough overlap. Consequently, the generated point clouds of the unwanted areas are of very low quality (i.e., flat treetops, missing object details, roofs directly connected to the ground, etc.). Figure 4 (a) shows the raw point clouds with the unwanted areas located at the boundaries. It is worth pointing out that the unwanted areas will not only reduce visual quality in the generated virtual environment but also add tremendous challenges for the segmentation task. Thus, removing the data in the unwanted areas is a necessary step.

Camera positions are used in this study to define the AOI. The camera positions are first projected onto the 2D plane (i.e., XY-plane). Following that, a 2D concave hull is computed, having all the camera positions inside. Points in the point cloud fall inside the concave hull are then extracted. Figure 4 (b) shows all image positions and Figure 4 (c) shows the cropped point cloud inside the AOI.

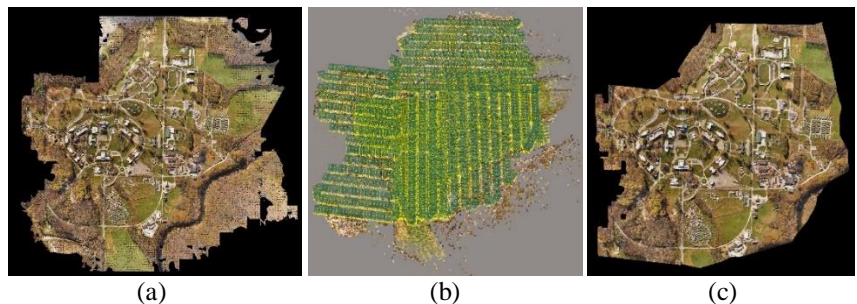

**Figure 4. Selecting Area of Interest (AOI).**

**Voxelization**

As previous study (Huang & You, 2016) pointed out that 3D voxel representation of a point cloud is suitable for many CNN based classification, segmentation, and object detection algorithms. Thus, the cleaned point cloud from the previous process is turned into 3D voxel grids. Given a point cloud, we first divide the 3D space into large voxels with a predefined size. The size of the large voxel is defined with its width ($W_L$), depth ($D_L$), and height ($H_L$) along X, Y, and Z axes respectively. Following that, for each large voxel that has points in it, we subdivide it into small voxels with another predefined size, the size of the small voxel is defined with its width ($W_S$), depth ($D_S$), and height ($H_S$) along X, Y, and Z axes respectively. Note that $W_L$, $D_L$, and $H_L$ need to be divisible by $W_S$, $D_S$, and $H_S$ respectively. This would result in a grid of size ($W_N \times D_N \times H_N$), which would be readily used for the deep convolutional neural network in the next section.

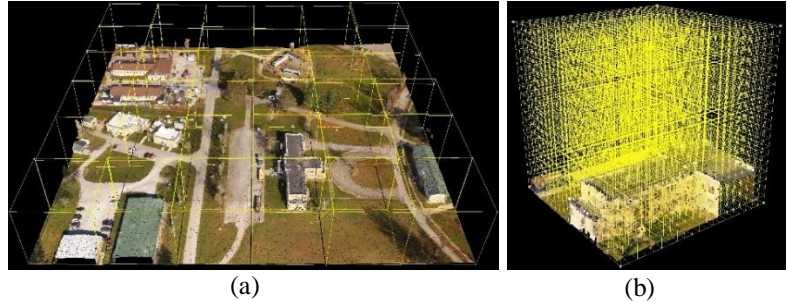

**Figure 5. Voxelization. (a) Large voxels. (b) Small voxels.**

Figure 5 (a) shows the partition of a point cloud into large voxels, and Figure 5 (b) shows the small voxels in a large voxel. The segmentation process is performed on each large voxel, and labels are assigned to all small voxels. It is worth pointing out that the size of voxels will affect the segmentation result. Larger voxel size will provide the segmentation model a bigger proximity of the nearby objects inside the scene and improve the segmentation result. However, large voxel size will slow down the training process and may even make the training process impractical. In this study, the size of large voxels was set to 40 m × 40 m × 40 m, and the size of small voxels was set to 0.5 m × 0.5 m × 0.5 m.

**U-Net**

For the voxel segmentation network, we adapted a network architecture similar to U-Net (Ronneberger, Fischer, & Brox, 2015). This method is a fully convolutional network that has been applied successfully for 2D image segmentation. The diagram in Figure 6 summarizes the network architecture that is used in our system. The input of the network is a voxel of size ($W_N \times D_N \times H_N$) with a single occupancy channel, where 1 indicates the cell is occupied and 0 otherwise. As shown in the diagram, the voxel is first fed through series convolution and max-pooling layers to extract the bottleneck feature map. The high-level feature map is then recursively up-sampled and concatenated with a feature map from the previous level to obtain the feature map in original voxel resolutions. Finally, a $1 \times 1 \times 1$ convolution layer is applied on the final feature map to assign a class label for each small voxel within the large voxel. To improve training stability and efficiency, we also add a batch normalization layer between each convolution layer (Ioffe & Szegedy, 2015).

Since we only use occupancy information and ignore the RGB colors in the input data to train the network, the resulting model relies on only scene geometry to determine the segmentation labels. Therefore, it should be able to generalize between regions with different color schemes. To further avoid overfitting, we also introduce a drop-out layer with a 50% drop-out rate after each max-pooling layer. This could cause some fluctuations of training loss during an epoch but overall would improve the accuracy when testing on an unseen dataset.

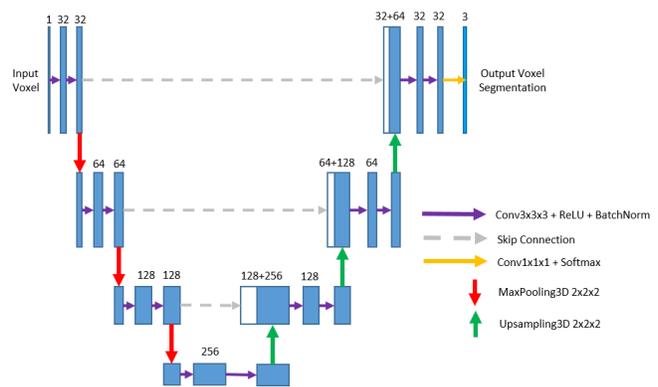

**Figure 6. U-Net Architecture Used in Our System**

**Database and experiments**

The designed top-level terrain elements segmentation workflow was tested using our collected datasets over the past 2 years. A total of 20 datasets were manually segmented into man-made structures, vegetation, and ground. Note that, grass areas are considered as ground instead of vegetation in this study. A U-Net model is trained on 19 datasets and tested on the MUTC dataset. The segmentation result is shown in Figure 7. One limitation of the designed workflow is that, there are cases where a group of points on roofs is mislabeled as ground. One possible solution is to design separate networks specialized on segmenting ground vs non-ground and man-made objects vs vegetation. This way, the mislabeled roof points can be identified using designed post processing and moved into the non-ground category and then segmented into man-made structures or vegetation with the second network.

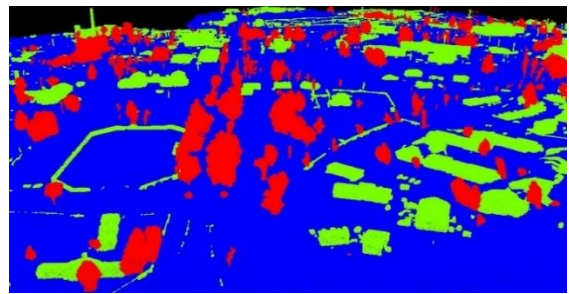

**Figure 7. Point Cloud Segmentation Result of MUTC**

A quantitative analysis was performed to compare the point cloud segmentation using the proposed STPLS+ (a fully automated process) and the authors previously developed STPLS (a semi-automated process) (Chen et al., 2019). MUTC dataset was used for the comparison. The result is summarized in Table 1.

It is worth pointing out that the data processing time for STPLS+ is purely computational. For STPLS, this processing time includes the time for creating the required training data manually, and this manual process highly depends on the user familiarity with the 3D annotation software. For a typical user familiar with STPLS system, it takes about 1 hours/km$^2$ of manual efforts and thus the previous system was not scalable to large terrains. STPLS+ was also able to improve the segmentation accuracy for ground and manmade structure since it is using the state-of-the-art deep learning technique for feature extraction whereas STPLS was using handcrafted features.

**Table 1. Example of a Single Column Table**

|  | STPLS+ | STPLS |
|---|---|---|
| **Creating training data manually** | 0% | 10% |
| **Data processing time** | 2 hours/km$^2$ | 1.5 – 3 hours/km$^2$ |
| **Ground segmentation accuracy** | 97% | 96% |
| **Manmade structure segmentation accuracy** | 87% | 86% |
| **Vegetation segmentation accuracy** | 93% | 93% |

**Tree Location Identification**

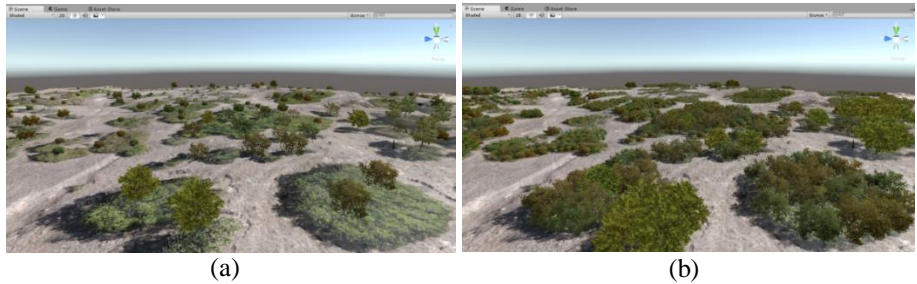

(a)       (b)

**Figure 8. Tree Replacement (a) The Old Tree Location Identification Approach; (b) The New Tree Location Identification Approach.**

The authors previously designed a two-step tree locations identification algorithm (Chen et al., 2019). A connected component algorithm was used to first cluster points into different rough clusters. Following that, K-means algorithm was applied to separate a connected rough cluster into multiple trees.

One drawback was that the algorithm would run multiple iterations with different parameters of K to find a suitable number of clusters. Moreover, users must provide a fixed distance parameter Dist, to determine the minimum spacing between adjacent tree locations. If the Dist is set to be large, then the tree locations would be sparse. This would result in less realistic tree placements in a bush area, as shown in Figure 8 (a) where only a few trees are placed in a dense bush area. To alleviate this issue, we propose to estimate the number of trees within a bush area based on the area of coverage and Poisson sampling. To estimate the area that is covered by the tree point cloud, we apply Delaunay triangulation to generate a 2D area spanned by the point clouds. Then based on the average heights of the points, we would estimate typical tree size and thus the number of trees that should be spanning the area.

Once we have the coverage area and the number of trees, the next step is to estimate actual tree locations within the area. While using K-means clustering would yield reasonable results, it sometimes produced uneven tree locations around the area with denser points. To produce more uniformly distributed tree locations, we applied Poisson-disc sampling within the tree area and used the sampled points as the cluster center. Figure 9 shows a comparison between the two methods. In comparison, K-means clustering is biased to generate more trees around the area with high point density, while Poisson-disc sampling would produce more uniform tree locations. With these improvements, the resulting tree placements look better especially in the bush area, as shown in Figure 8 (b).

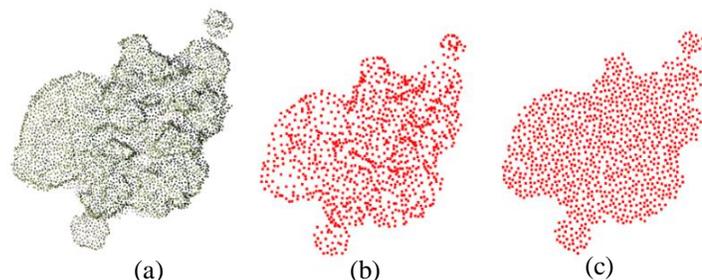

(a)       (b)       (c)

**Figure 9. Tree Location Identification (a) Original Tree Point Clouds (b) K-means Clustering (c) Poisson-disc Sampling**

**Ground Material Classification**

Recognizing surface materials from 2D images is a fundamental problem in computer vision and has drawn attention from both academia and industry in the past two decades. The problem is usually viewed as a texture classification problem and has been studied side by side with image classification and object detection problems. Earlier works have been focused on investigating and developing statistical approaches to quantify handcrafted texture features that were extracted using different texture filters (i.e., Sobel filter, color histograms, filter banks, Gabor filters, etc.) for the material classification problems (Varma & Zisserman, 2009). However, these statistical approaches could only achieve a classification accuracy of 23.8% when applied to a database that contains common materials with real-world appearances (i.e., Flickr Material Database - FMD (Sharan, Rosenholtz, & Adelson, 2009)).

With the recent advancement in deep learning, researchers started to investigate and develop different CNN architectures for solving the material classification tasks. M. Cimpoi et al. extracted texture features using VGG model and performed the classification using SVM, achieving a 82.4% classification accuracy on FMD (Cimpoi, Maji, & Vedaldi, 2015). Previous studies have focused on and made valuable contributions to material classification for real-world images. However, the images that are rendered from photogrammetric-generated meshes (i.e., orthophotos) have a lower quality than the real-world images. The orthophotos could be distorted and blurred.

Furthermore, a thorough search of the relevant literature does not yield any existing material database that contains mesh rendered images. Thus, in this study, a ground material database was created using the photogrammetric generated orthophotos from USC, Fort Drum Army Base, The Camp Pendleton Infantry Immersion Trainer (IIT), and 29 palms range 400.

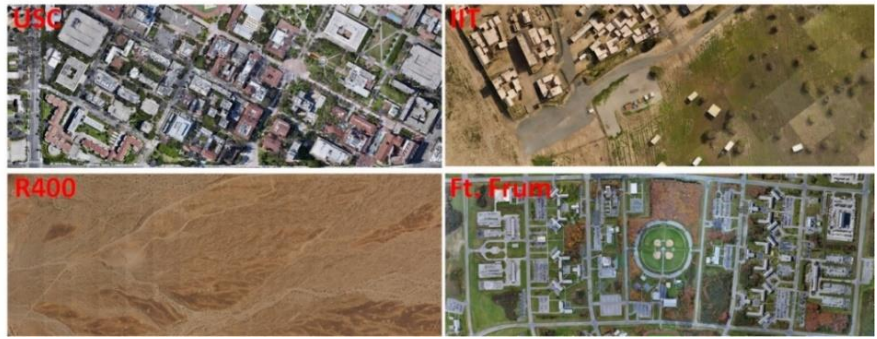

**Figure 10. Ground Material Database.**

The classes used for ground material classification include (1) bare soil (2) road and (3) vegetation. Figure 10 shows the rendered orthophotos for the above-mentioned data sets. The process for creating the ground material database is shown in Figure 11. The rendered orthophotos were manually segmented into the predefined classes. Following that, the segmented orthophotos were cropped into small image patches. A class label was then assigned to each of the cropped image patches based on the amount of different ground materials the image patch contained. The assigned label was the ground material that the majority of the pixels belonged to. The created ground material database contains roughly 18,000 image patches (i.e., 6,000 image patches in each of the predefined classes).

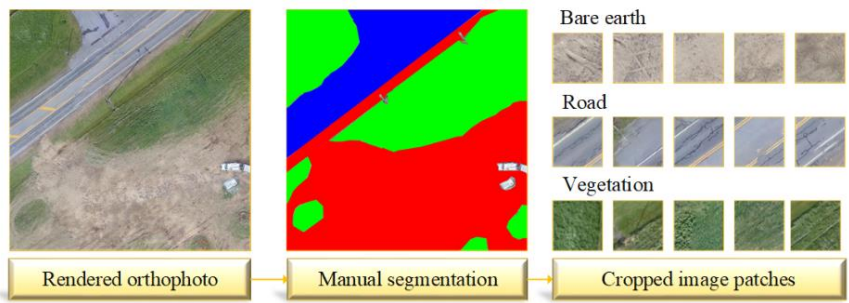

**Figure 11. Process for Creating Ground Material Database**

GoogleNet architecture (Szegedy et al., 2015) was selected as the neural network in this study. The classifier was trained on the above-mentioned database and tested on the MUTC dataset. The time for training the CNN classifier was 12 hours. The classification result was stored in a vector map in which each point represents the center of each cropped image patch. It is worth pointing out here that although the resolution of a computed-ground material vector map depends on the distance between two adjacent cropped image patches, it does not equal the size of the area covered by each cropped image patch. In this study, the distance between the two adjacent cropped image patches was set to 3 meters and the 5 by 5 meters area was covered by each cropped image patch. An experiment was conducted to evaluate the ground material classification approach with and without a fine-tuning process. Fine-tuning is the process that re-trains the fully connected layers in the CNN model using a small set of manually classified image patches from the testing data set. Five percent of the rendered orthophotos from MUTC dataset were manually segmented and used for the fine-tuning process.

The generated ground material vector map of the MUTC without and with the fine-tuning process is shown in Figure 12 (b) and (c), respectively. Bare soil, road, and vegetation are shown as blue, green and red, respectively. The time taken for the fine-tuning process was 13 minutes, and for the classification process was 15 minutes in both cases.

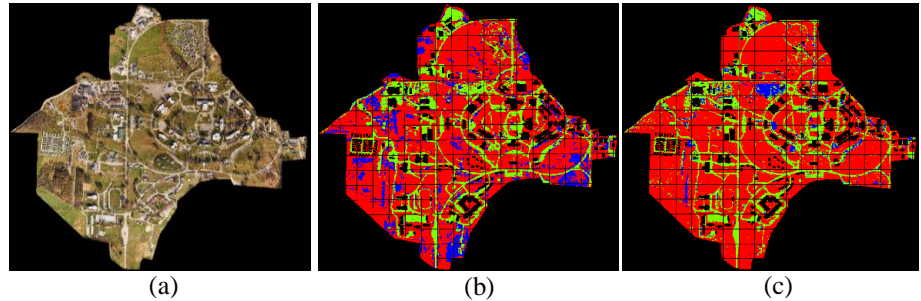

**Figure 12. Ground Material Classification Result (a) MUTC Dataset (b) Result without Fine Tuning (c) Result with Fine Tuning.**

## DATA IMPORT AND SIMULATION

As discussed earlier, meshes are needed in game engines for simulation. The photogrammetric-generated meshes are segmented based on the point cloud segmentation results since they are in the same coordinate system. An effective mesh segmentation process was designed in our previous study (Chen et al., 2019). Meshes are segmented by using the nearest neighbor algorithm to keep a set of vertices that are close enough to points in the segmented point cloud. Edges in the mesh will be kept, if both of its linking vertices are kept and will be eliminated if one of its linking vertices is excluded. It is worth pointing out that instead of removing tree meshes, we flatten them to the ground elevation so that there will be no holes on the ground when we replace the tree meshes with SpeedTree models. The produced segmented meshes are in a common mesh format – i.e., OBJ that can be imported into several different game engines (i.e., VBS, Unreal, Unity).

The authors have previously designed and developed a simulation tool — i.e., Aerial Terrain Line-of-sight Analysis System (ATLAS) using Unity game engines. The process for creating a virtual environment in ATLAS using the segmented meshes and extracted object information are discussed in the following paragraphs. Readers can refer to (Liles, Ortiz, & Caruthers, 2019) for more information about importing 3D terrain data for game and simulation engines.

Given the tree locations estimated in the previous section, we apply a modelization process to replace noisy photogrammetric tree meshes with higher quality geo-typical models. We first produce tree models of typical tree species and their seasonal varieties using SpeedTree software. The next step is to build an attribute table for these new trees, and therefore we could utilize such attributes to select the most similar tree model at each tree location during the tree replacement process.

Figure 13 shows some example of tree models and their attributes. Different attributes could be utilized such as tree crown shapes, tree heights, colors, and growth regions, etc. For our process, we select tree height and average leaf colors to measure the tree similarities. To avoid inappropriate tree types getting selected (i.e. a palm tree in Montana), we also integrate weather hardiness zones in the tree attributes by assigning a hardiness zone range for each tree model. Then based on the latitude and longitude of the map, we would exclude trees that are not suitable for the region during the tree replacement process.

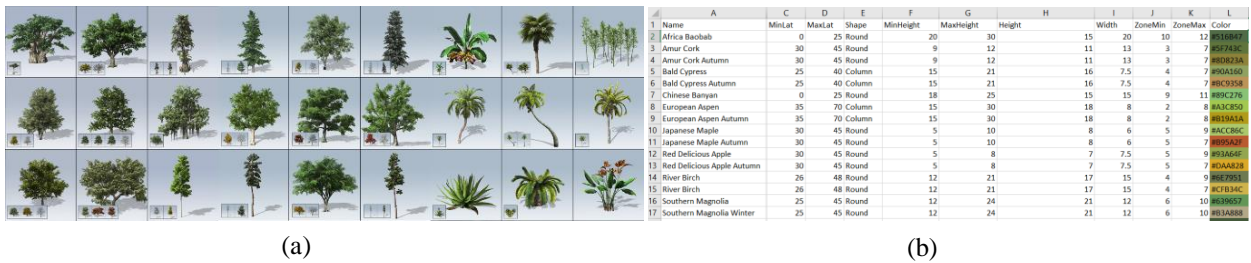

**Figure 13. (a) Tree models used for tree replacement process. (b) The tree attribute table built to guide the tree selection process at run-time.**

Figure 14 shows examples of tree replacement results for two different geographical locations. By following the weather hardiness constraints, vastly different tree species are selected for different regions. We also integrate the ground material classification results in the run-time simulation. Here, the material properties are used to adjust the edge weights of navigation mesh for pathfinding. the original navigation mesh assigns uniform weighting for each node and thus the A* pathfinding algorithm would simply search for the shortest geodesic path within the graph. By integrating ground material properties, we could assign various weights to the graph node based on its nearby materials and thus produce different pathfinding behaviors.

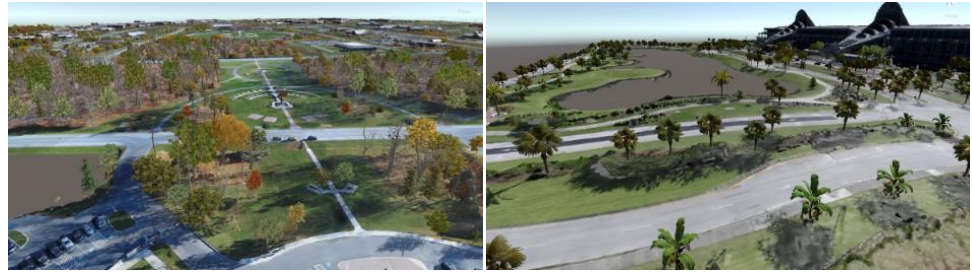

(a) (b)

**Figure 14. Examples of our tree replacement results. (a) Ft. Drum in New York state. (b) Orlando convention center in Florida.**

Figure 15 (a) and (b) show an example of the shortest path computed between the blue unit to its destination without and with using the ground material vector map. Edge weights for roads were set to a low value (0.2) and the edge weights for bare soil and vegetation were set to a high value (1.0) in the navigation mesh. This example illustrates that with the ground material classification process, the vehicle trafficability can be considered during the pathfinding simulation in a created virtual environment.

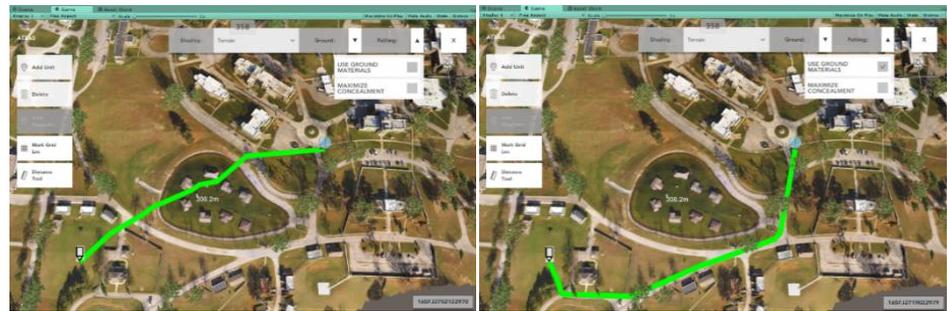

(a) (b)

**Figure 15. Path Finding Result Without and With Ground Material Classification.**

**APPLICATIONS & FUTURE WORK**

The implications of this research affect a large number of currently existing experimentation and development programs, and can additionally help shape new technologies and initiatives in ways that wouldn't be previously possible due to the quality and speed of our process. Inherent value in recognizing an object and understanding its physical attributes in a 3D space provide advanced knowledge not just for human intelligence but also for machine intelligence, resulting in higher quality ground truth for both simulation environments and ML model training. The future work will also seek to rapidly increase the fidelity in our own data sets, by using crowdsourced human ground truth annotations through Mechanical Turk, to collect human quality labeled data with which we can further train and validate our models, resulting in optimizations at all stages of our process. Additionally, we are working to improve aperture detection so that doors and windows would be properly identified and mensurated for a number of additional applications.

**Simulation Fidelity and Realism**

A notable example of current research benefitting from automated classification/segmentation is the increase of simulation realism and fidelity. As the classification/segmentation process expands to include a larger number of material components, every aspect of a simulation becomes more aligned with reality. The improved fidelity allows for accurate simulation and path finding using the two classification results achieved in this work to enhance the machine's understanding of the photogrammetric meshes.

First, using vegetative classification, the replacement of photogrammetric trees with size analogous geo-typical and geo-referenced trees. This improves the quality of both human and machine understanding as to what paths are possible since photogrammetric trees with current technology are subject to a number of processes in the pipeline that degrade their fidelity, among these are natural occurrence such as dynamic wind direction among multiple source photos, and artificially generative occurrences such as the limitations of photogrammetric point cloud mesh creation.

Secondly, improvement in fidelity for pathfinding comes from ground material classification, allowing the machine to understand what ground surface would produce the easiest traversal through weighted penalties in an A* Path Graph. This improved understanding of terrain and especially trafficability can be put into immediate use with human intervention, but in the future would allow an Automated Command and Control system to instantly calculate and classify all components of a battlespace, beyond solely optimized paths returning valid data for all METT-TC considerations. It is also worth pointing out that the idea of replacing low quality meshes with geo-typical models are not limited to vegetation. Other mesh objects such as light poles, cars, and buildings can also be replaced if the location and the appropriate attributes can be extracted. For instance, building meshes can be replaced with geo-typical building if the building footprint can be extracted along with the building height, roof styles, and window locations. The authors are planning to investigate and develop approaches for extracting these objects in our future works.

**AI Training**

The machine understanding of improved ground truth fidelity also enables a rapid increase in the speed at which AI entities can be trained in 3D environments. In our research efforts, we improved the classification learning algorithms by providing additional ground truth and training data specific to our intended applications. These improvements in turn improve the fidelity of data that is provided to subsequent AI entities who interact and learn within the space itself. In this way, we can determine how well the initial classification algorithms and heuristics have done their job, not only by our own reviews, but also by analyzing and understanding any assumptions made by AI entities "living" in the space.

For instance, if we deploy an AI infantry unit into a 3D simulation where we have provided extended attribution for ground materials, after interacting with such high fidelity ground truth, the agent should be able to make better probabilistic determinations in any additional environment unknown to it due to the quality of the output from the original classification algorithms. Conversely, any areas that had confused this entity would highlight the need for additional training or modification to the original deep learning classification process. This would create a cycle of improvements that would increase the fidelity and intelligence of both the original process and results. This improved fidelity becomes especially meaningful when it comes to lowering the computational burden of very large-scale autonomous simulations.

The quality of the ground-truth can be used directly by hundreds of thousands of agents through direct interaction and pre-calculated navigation meshes. This process is currently being researched by the One World Terrain project with the experimental "Data Oriented Tech Stack" of the Unity Game Engine (McCullough, New, Nam, & McAlinden, 2019).

**Environment Destructibility and Deformation of 3D Meshes**

For realistic simulation, 3D visualized terrain data should be dynamically modifiable with the closest precision to reality regardless of its original source form or pre-existing attribution. Using real-world understanding and input to address simulated terrain analysis and visualization over time from common effect, including total/partial structural damage and terrain deformation both natural and man-made (i.e. blast/impact damage and cratering, fires, earthquakes, trenches, mounds, etc.) a full model of understanding can be created to accurately predict and plan for the outcome of any number of affecting circumstances. However, in order to accurately predict and produce this model, material classification is necessary to properly assign the needed parameters to their corresponding meshes.

For instance, a blast over sand will have a significantly different effect as opposed to a blast over concrete. Not only will the sand propagate differently, but the depression caused via deformation would be more significant, whereas concrete would fracture and depending on the size of the blast may not deform or depress the meshes beyond that. Additionally, in a battlespace where it is necessary to know the effect of explosive ordnance with regard to breaching, wood will be affected much differently than steel.

**CONCLUSION**

In this paper, we have presented our designed fully automated approach for segmenting photogrammetric-generated point cloud into top-level terrain elements (i.e., man-made structure, vegetation, and ground). We have created a point cloud database which contains 20 annotated datasets. The segmentation approach was validated by training on 19 datasets and tested on the 20th dataset. The individual tree locations were extracted from the segmented vegetation point cloud. An improved tree location identification approach was designed based on the authors' previous work. The new approach works well on bush areas as shown in the testing result. Tree types were assigned to each tree location by using the extracted tree attributes (i.e., crown shapes, tree heights, colors, and growth regions) and the weather hardiness zones.

We also presented a process to classify ground materials using orthophotos. A ground material database was created by annotating and cropping orthophotos from our previous collects. The database contains 18,000 cropped image patches. A ground material classification network based on GoogleNet was trained using the database and tested on the MUTC dataset. The results showed that the trained model can be applied directly for a new dataset and produce reasonable result, noise can be reduced if a fine-tuning process is performed. This paper also highlighted several potential applications of using the segmented and the extracted data for training and simulation.

## ACKNOWLEDGEMENTS

The authors would like to thank the two primary sponsors of this research: Army Futures Command (AFC) Synthetic Training Environment (STE), and the Office of Naval Research (ONR). They would also like to acknowledge the assistance provided by 3/7 Special Forces Group (SFG), Naval Special Warfare (NSW), the National Training Center (NTC), and the US Marine Corps. This work is supported by University Affiliated Research Center (UARC) award W911NF-14-D-0005. Statements and opinions expressed and content included do not necessarily reflect the position or the policy of the Government, and no official endorsement should be inferred.